\documentclass[letterpaper]{article} % DO NOT CHANGE THIS
\usepackage{aaai25}  % DO NOT CHANGE THIS
\usepackage{times}  % DO NOT CHANGE THIS
\usepackage{helvet}  % DO NOT CHANGE THIS
\usepackage{courier}  % DO NOT CHANGE THIS
\usepackage[hyphens]{url}  % DO NOT CHANGE THIS
\usepackage{graphicx} % DO NOT CHANGE THIS
\usepackage{natbib}  % DO NOT CHANGE THIS AND DO NOT ADD ANY OPTIONS TO IT
\usepackage{caption} % DO NOT CHANGE THIS AND DO NOT ADD ANY OPTIONS TO IT
\usepackage{subcaption}
\usepackage{microtype}
\usepackage{url}
\usepackage{booktabs}

\usepackage{times}
\usepackage{longtable}
 \usepackage{latexsym}

\usepackage{amsmath}
\usepackage{adjustbox}
\usepackage[utf8]{inputenc} % allow utf-8 input
\usepackage[T1]{fontenc}    % use 8-bit T1 fonts   % hyperlinks
\usepackage{booktabs}       % professional-quality tables
\usepackage{amsfonts}       % blackboard math symbols
\usepackage{nicefrac}       % compact symbols for 1/2, etc.
\usepackage{microtype}      % microtypography
\usepackage{xcolor}         % colors
\usepackage{multirow}
\usepackage{graphicx}
\usepackage{caption}

\title{Unveiling the Impact of Coding Data Instruction Fine-Tuning on Large Language Models Reasoning}

\author{Xinlu Zhang$^1$\thanks{Corresponding Author: xinluzhang@ucsb.edu} , 
Zhiyu Zoey Chen$^2$, Xi Ye$^3$, Xianjun Yang$^1$, Lichang Chen$^4$, \\ \textbf{William Yang Wang}$^1$, \textbf{Linda Ruth Petzold} $^1$ }
\affiliations{
\textsuperscript{\rm 1}University of California, Santa Barbara \\
\textsuperscript{\rm 2}The University of Texas at Dallas\\
\textsuperscript{\rm 3}The University of Texas at Austin\\
\textsuperscript{\rm 4}University of Maryland, College Park
} 

\usepackage{bibentry}
\begin{document}
\maketitle

\begin{abstract} 
Instruction Fine-Tuning (IFT) significantly enhances the zero-shot capabilities of pretrained Large Language Models (LLMs). While coding data is known to boost LLM reasoning abilities during pretraining, its role in activating internal reasoning capacities during IFT remains understudied. This paper investigates a key question: \textit{How does coding data impact LLMs' reasoning capacities during IFT stage?} To explore this, we thoroughly examine the impact of coding data across different coding data proportions, model families, sizes, and reasoning domains, from various perspectives. Specifically, we create three IFT datasets with increasing coding data proportions, fine-tune six LLM backbones across different families and scales on these datasets, evaluate the tuned models' performance across twelve tasks in three reasoning domains, and analyze the outcomes from three broad-to-granular perspectives: overall, domain-level, and task-specific.
Our holistic analysis provides valuable insights into each perspective. First, coding data tuning enhances the overall reasoning capabilities of LLMs across different model families and scales. Moreover, while the impact of coding data varies by domain, it shows consistent trends within each domain across different model families and scales. Additionally, coding data generally provides comparable task-specific benefits across model families, with optimal proportions in IFT datasets being task-dependent.

\end{abstract}
\vspace{-2mm}
\section{Introduction}
\vspace{-1mm}
Large Language Models (LLMs) have significantly advanced in task generalization by training on diverse text data sources \citep{touvron2023llama,touvron2023llama-2,brown2020language,jiang2023mistral} and instruction-finetuning (IFT) further elicits their intrinsic abilities in a zero-shot manner \citep{ouyang2022training, longpre2023flan, wei2022finetuned,openai2022chatgpt,openai2023gpt4,claude-2}. Although previous studies \citep{peng2023instruction, alpaca, xu2023wizardlm,vicuna2023} have improved IFT dataset diversity to better align LLMs with human needs, the impact of specific data types during IFT remains underexplored.

\begin{figure*}[t!]
\centering
\includegraphics[width=0.85\textwidth]{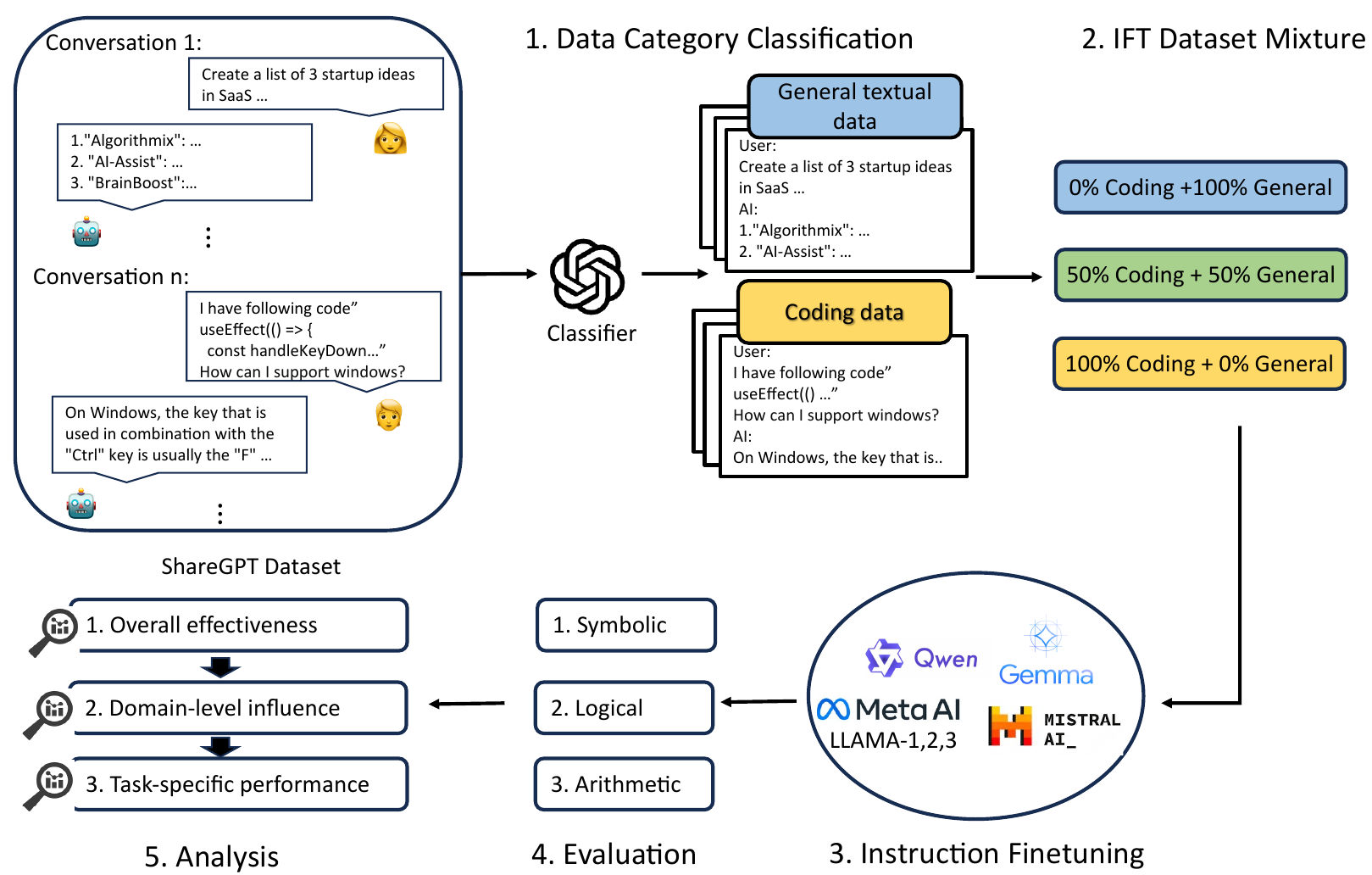}
\vspace{-3mm}
\caption{ \textbf{Pipeline Overview.} The process utilizes ShareGPT data as a starting point. \textbf{1. Data Category Classification:} Using ChatGPT to classify code instances within ShareGPT to obtain a code-centric IFT dataset. \textbf{2. IFT Data Mixture:} Constructing three IFT mixture datasets with increasing proportions of coding data. \textbf{3. Instruction Finetuning:} Fine-tuning LLMs from six families across various scales with these three IFT datasets, respectively.  \textbf{4. Evaluation:} Evaluating the fine-tuned models' reasoning capacities across three domains. \textbf{5. Analysis:} Analyzing from three broad-to-granular perspectives.} 
\label{fig: pipeline}
\vspace{-5mm}
\end{figure*}

Coding data, with its logical consistency and reduced ambiguity compared to natural text, has been empirically shown to enhance LLM reasoning capabilities during pretraining \citep{liang2023holistic,fu2022gptroadmap, ma2023training, guo2024deepseekcoder}. This enables LLMs to acquire advanced intrinsic knowledge, supporting complex reasoning in real-world applications like text summarization \citep{yang2023exploring}, numerical problem solving \citep{luo2023wizardmath, yue2023mammoth}, and knowledge-intensive tasks \citep{chen2024survey}. However, at the IFT stage, research primarily shows that coding data tuning improves coding-related in-domain performance \citep{yuan2023evaluating, luo2023wizardcoder, ma2023training}. The impact on \textbf{out-of-domain general reasoning capabilities} remains underexplored due to complex variations in coding data proportions, model backbones, and reasoning task types \citep{Wei2022ChainOT, ma2023training, liang2023holistic}. Therefore, we raise a natural question: \textit{How does coding data impact LLMs' reasoning capacities during the IFT stage?}

To thoroughly answer this question, we propose an analysis pipeline to investigate how coding data affects LLMs' reasoning capacities during IFT, considering coding data proportions, model families, scales, and reasoning domains. We create IFT datasets from ShareGPT \citep{sharegpt} using ChatGPT \citep{openai2022chatgpt} to classify instances as either coding or general text, forming code-centric and general textual datasets. We then generate three IFT datasets with coding data proportions of 0\%, 50\%, and 100\%, maintaining consistent dataset sizes. Six base models of varying families and scales —Llama-1 \citep{touvron2023llama}, Llama-2 \citep{touvron2023llama-2}, Llama-3 \citep{llama3modelcard}, Mistral \citep{jiang2023mistral}, Qwen-1.5 \citep{qwen}, and Gemma \citep{team2024gemma}—are fine-tuned on these datasets. We evaluate the tuned models on symbolic, logical, and arithmetic reasoning domains. Finally, we analyze coding data impact from three perspectives: overall effectiveness, domain-level influence, and task-specific performance, as shown in Figure \ref{fig: pipeline}.

To our best knowledge, this is the first work to thoroughly analyze how coding data affects LLMs' reasoning capacities during IFT, across different coding data proportions, model families, scales, and reasoning types. We gain significant insights and summarize the main findings for each perspective.

\noindent \textbf{Overall effectiveness.} As the proportion of coding data for tuning increases, we observe consistent and gradual performance enhancements across different model families and scales. However, improvements vary among model backbones. Compared to tuning on the pure natural text dataset, the greatest overall improvement is achieved on Mistral with a 10.3 percentage points gain, while a more modest 1.5 percentage points gain obtained on Llama-3. Analysis of the models' responses indicates that coding data tuning enhances overall reasoning capacities for problem-solving, extending beyond in-domain programming skills.

\noindent \textbf{Domain-level influence.} Coding data tuning elicits different reasoning abilities in varied ways. There is marked improvement in the symbolic domain, which includes foundational reasoning skills. However, in arithmetic reasoning, which involves real-world math problems, performance gaps appear compared to models tuned on general textual datasets, addressing more diverse human needs. Additionally, we observe consistent performance trends across various model backbones and sizes within each reasoning domain, indicating the potential transferability of coding data effects during the IFT stage. Further analysis shows that models tuned with coding data can more adeptly apply appropriate skills for solving questions based on different domain properties.

\noindent \textbf{Task-specific performance.} Coding data typically yields comparable task-specific benefits across different model families, with a similar number of tasks showing improvement in two out of three reasoning domains. However, obtaining optimizing strategies for mixing coding and natural textual data presents a challenge. While the majority of optimal coding data proportions for improving task performance remain consistent across model families, there is no single coding data proportion setting that consistently enhances task-specific reasoning abilities better than another.

\vspace{-2mm}
\section{Related Work}
\vspace{-2mm}
\noindent\textbf{IFT. }IFT has proven effective in enhancing pretrained LLMs for zero-shot tasks \citep{ouyang2022training, longpre2023flan}. \citet{wei2022finetuned, chung2022scaling} used IFT datasets from NLP benchmarks, improving generalization but often not fully aligning with real-world user intentions due to simpler instructions. In contrast, \citet{ouyang2022training} developed InstructGPT by tuning on a diverse dataset of real-world instructions and responses, better meeting user needs. Open-source models \citep{alpaca, xu2023wizardlm, vicuna2023, peng2023instruction,zhang2024alpacareinstructiontuned} fine-tuned on diverse IFT datasets from strong teacher models have shown that such diversity enhances alignment with complex user intents \citep{chen2023alpagasus}. However, the impact of different IFT data types on LLM effectiveness remains underexplored. This work studies the effect of coding data, known for logical clarity and structure, during IFT.

\noindent\textbf{Reasoning in LLMs.}  Reasoning involves logically analyzing a subject, using evidence and prior knowledge to reach conclusions \citep{wason1972psychology,wason1968reasoning}. It has been seen as one of LLMs' emergent behaviors, shown as models are large enough \citep{wei2022emergent,Wei2022ChainOT}. Although improving models' reasoning capacities shows promising results in different applications \citep{li2022explanations,zhang2023enhancing}, comprehensively evaluating these capacities remains challenging due to their complex nature, requiring different fine-grained abilities within different subdomains \citep{Wei2022ChainOT, ma2023training,liang2023holistic,qiu2023phenomenal}. We aim to thoroughly evaluate models across the subdomains of symbolic, logical, and arithmetic reasoning to deepen understanding of LLMs' reasoning capabilities at the IFT stage.

\noindent\textbf{Code in LLMs. } Integrating code into LLMs enhances their performance in programming, complex reasoning, and structural knowledge capture \citep{ma2023training, liang2023holistic, fu2022gptroadmap, wang2023code4struct, madaan2022language}. Most investigations focus on the pretraining stage. \citet{ma2023training} shows that LLMs pretrained with coding data outperform those trained with only natural language in both code-related and general tasks. \citet{liang2023holistic} demonstrates that Codex excels in complex mathematical reasoning, and \citet{madaan2022language} highlights coding data’s role in improving structural reasoning. During the IFT stage, \citet{ma2023training} reveals that coding data enhances in-domain abilities. Conversely, we study how coding data elicits out-of-domain reasoning capacities in pretrained LLMs.

\vspace{-2mm}
\section{Experimental Setting} \label{sec: expsetting}
\subsection{IFT Data Construction} \label{sec:data}
\vspace{-1mm}
We use ShareGPT \citep{sharegpt}, which collects real-world user inquiries, as our data source. After deduplication and extraction of the initial round of Human-AI conversations, we obtain a dataset of 45,742 instances. Using GPT-3.5-turbo \citep{openai2022chatgpt}, we categorize these conversations into three groups: \texttt{Code}, \texttt{Math}, and \texttt{Others}. Conversations involving coding data are classified as \texttt{Code}, those related to mathematical concepts and problems as \texttt{Math}, and all other general natural language texts as \texttt{Others}. The categorization results in 10,196 \texttt{Code}, 1,481 \texttt{Math}, and 34,065 \texttt{Others} instances. We exclude the \texttt{Math} category to prevent its potential influence on the model's reasoning capabilities during tuning. For further experimentation, we select a random subset from the \texttt{Others} category, termed \texttt{General}, equal in size to the \texttt{Code} category, to facilitate fair analysis. Additionally, we establish a \texttt{Half-half} setting by mixing equal portions of data from the \texttt{Code} and \texttt{General} categories. This setup produces three equal-size IFT datasets: \texttt{General}, \texttt{Half-half}, and \texttt{Code}, containing 0\%, 50\%, and 100\% coding data, respectively. This arrangement enables us to explore how increasing the proportion of coding data impacts LLMs' reasoning abilities in the IFT stage. The detailed categorization prompts for ShareGPT and coding data are in Table \ref{tab:prompt-classfication} and Table \ref{tab:prompt-datadiversity} in Appendix \ref{subsec:prompt}, and the corresponding code category diversity analysis is in Appendix \ref{subsec: code_diversity}.

\vspace{-2mm}
\subsection{Task Description}
\vspace{-1mm}
To thoroughly assess the models' reasoning capabilities, we evaluate them on twelve generative tasks, across three reasoning types: symbolic, logical, and arithmetic. \textbf{Symbolic}: We focus on four tasks \citep{Wei2022ChainOT}: (1) First Letter Concatenation, (2) Last Letter Concatenation, (3) Reverse List, and (4) Coin Flip. \textbf{Logical}: We utilize four tasks, requiring strong logical ability: (1) Cluttr \citep{cluttr}, (2) List Functions \citep{rule2020child}, (3) Babi-Induction and (4) Babi-Deduction \citep{weston2015towards}. \textbf{Arithmetic}: Four arithmetic benchmarks are involved to evaluate the mathematics world problem-solving ability (1) GSM8K \citep{cobbe2021training}, (2) SVAMP \citep{patel2021nlp}, (3) ASDiv \citep{miao-etal-2020-diverse}, and (4) MAWPS \citep{mawps}.
For symbolic reasoning, we generate synthetic datasets following \citet{fortes2023simple} and balance the representation of difficulty levels for each task. For example, we generate 500 instances for names containing 2 to 4 words in letter concatenation tasks. For other tasks, we evaluate the models with the test sets for each task when publicly available. Otherwise, the development sets are used. Details on data statistics, the reasoning domain selection and synthetic datasets generation are in the Appendix \ref{sec: reasoningtask}.

\vspace{-2mm}
\subsection{Instruction Fine-tuned LLMs}
\vspace{-1mm}
To systematically assess the coding data impact, we conduct experiments with six distinct LLM families: Llama-1 \citep{touvron2023llama}, Llama-2\citep{touvron2023llama-2}, Llama-3 \citep{llama3modelcard}, Mistral-v0.1 \citep{jiang2023mistral}, Qwen-1.5 \citep{qwen}, and Gemma \citep{team2024gemma}. Each model is fine-tuned on the three uniquely composed IFT datasets—\texttt{General}, \texttt{Half-half}, and \texttt{Code}, respectively. Each fine-tuned model is evaluated on the twelve reasoning tasks in three reasoning domains. The training prompt is deferred in hyperparameter settings of each model are available in the Appendix \ref{sec:hyp}. 
\vspace{-2mm}
\subsection{Evaluation Setup}
\vspace{-1mm}
Our evaluation operates in a \textit{zero-shot} setting, where models are prompted to generate responses to corresponding questions without additional context, minimizing external influences. We standardize experimental conditions by limiting the maximum token length to 1024 and employing greedy decoding for all model outputs. Given the variability in output styles, we utilize GPT-3.5-turbo \citep{openai2022chatgpt} as an extractor to parse predictions from the generated text. These predictions are subsequently compared to the ground truth using accuracy (\%) as the evaluation metric. The generation prompt for each dataset and answer extraction prompt is provided in Table \ref{tab:prompt-generation} and Table \ref{tab:prompt-extractor} of the Appendix.

\begin{table*}[t!]
\centering
\caption{\textbf{Overall reasoning comparison of models tuned on \texttt{General}, \texttt{Half-half}, and \texttt{Code} with increasing coding data proportion, across different families.} Results show average scores (\%) across 12 reasoning tasks. The \textbf{Best} setting per model family is in bold. $\Delta$ and $\eta$ indicate absolute and relative gains, respectively, between the best-performing coding data setting and \texttt{General}.}
\vspace{-4mm}
\resizebox{0.85\linewidth}{!}{
\small
\begin{tabular}{c  cccc c c}
\toprule
Coding data prop.(\%)& Llama-1-7B & Llama-2-7B & Llama-3-8B &  Mistral-7B-v0.1 & Qwen-1.5-7B & Gemma-7B\\
\midrule
\texttt{General} (0\%)  & 23.2 & 26.7& 41.4 & 42.0 & 42.2& 22.3\\
\texttt{Half-half} (50\%) & 25.0& 29.2 & \textbf{42.9} &43.9 & 44.9 & 23.7\\
\texttt{Code} (100\%) & \textbf{27.9} & \textbf{30.6} & 42.6 & \textbf{52.3}  & \textbf{47.4} & \textbf{30.0} \\
\midrule
$\Delta$ &+4.6 &+3.9 & +1.5 & +10.3 &+ 5.5 &+ 7.7\\
$\eta$ &+20.0\% & +14.5\% & +3.6\% &+24.6\% &  +13.0\%& +34.7\%\\
\bottomrule
\end{tabular}
}
  \vspace{-2mm}
\label{tab:7bperformance}
\end{table*}

\begin{figure*}[t!] % Add [t] option for 
  \begin{minipage}{0.50\textwidth}    \includegraphics[width=\linewidth,height=0.45\linewidth]{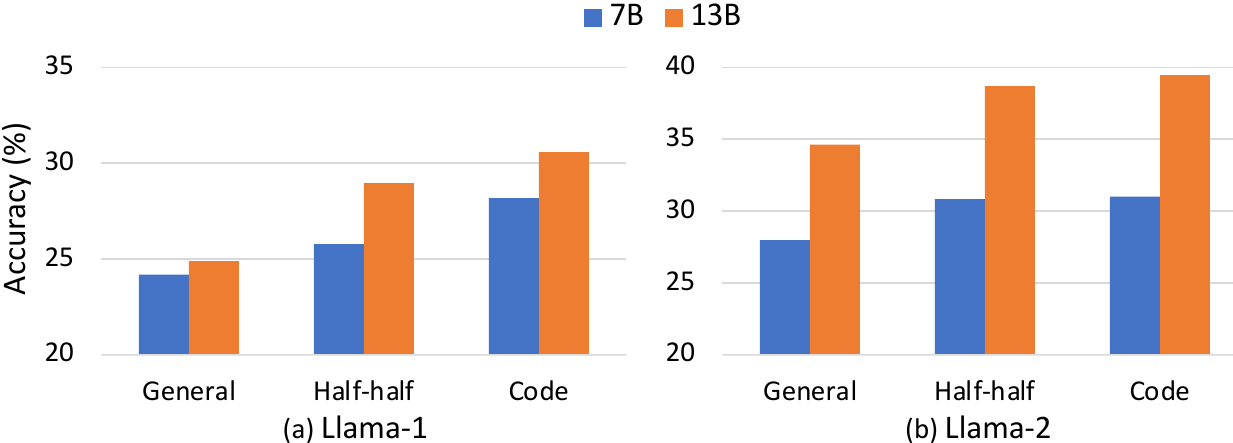}
  	\centering 
   \vspace{-6mm}
    \caption{\textbf{Overall comparison across different model scales.} Results show the overall performance of models using Llama-1 and Llama-2 with 7B and 13B parameters as backbones.}    
    \label{fig:modelscale}
  \end{minipage}
    \hfill
    \begin{minipage}{0.43\textwidth}
\includegraphics[width=\linewidth,height=0.6\linewidth]{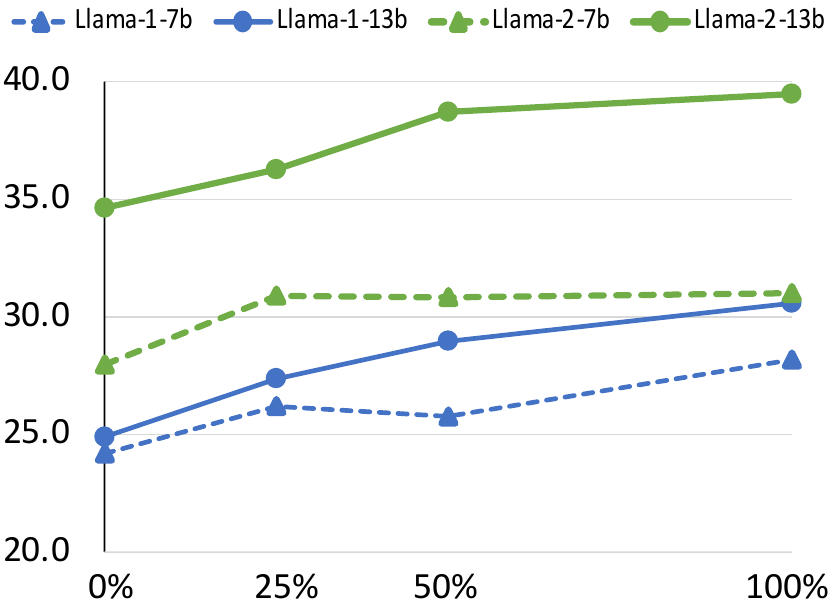}
	\centering 
 \vspace{-6mm}
    \caption{\textbf{Overall performance comparison across different coding data percentages.} }
    \label{fig:modelspropotion}
  \end{minipage}
\end{figure*}

\vspace{-2mm}
\section{Experimental Results} \label{sec:exp}
\subsection{Results on Overall Performance} \label{sec:overall}
\vspace{-1mm}
We first compare the average results of 12 reasoning tasks between models tuned with different coding data proportions across 6 LLM families. Results are shown in Table \ref{tab:7bperformance}.

\noindent\textbf{Coding data successfully elicits LLM reasoning capacities in IFT.} 
Overall, we observed a gradual improvement in average accuracy as the proportion of coding data increased during the IFT stage. These consistent gains across different model families clearly demonstrate the benefits of the specialized knowledge that coding data provides, effectively enhancing models' reasoning capabilities. 
Notably, for using Mistral-7B-v0.1 as the base model, tuning with \texttt{Code}, achieves a significant absolute performance gain of 10.3 compared to tuning with \texttt{General}. This underscores that datasets rich in code are crucial for eliciting the advanced reasoning abilities of LLMs.

\noindent\textbf{The improvements brought by coding data are divergent across model families.} Although different model families show positive effects from tuning with coding data during IFT, the improvements vary across model families. For example, the Mistral-7B-v0.1 and Gemma-7B exhibit substantial benefits, achieving 24.6\% and 34.7\% relative gains, respectively, when fine-tuning with \texttt{Code} compared to \texttt{General}. Conversely, the Llama-3-8B model shows a more modest improvement, achieving a 3.6\% relative gain. This disparity could be due to differences in the LLMs pretraining, during which models primarily acquire intrinsic knowledge, rather than the IFT stage \citep{albalak2024survey}.

\noindent\textbf{Coding data achieves consistent gains as the LLM scales up.} To examine the impact of coding data as LLM size increases, we tune Llama-1 and Llama-2 with 7B and 13B parameters under three coding proportion settings. The results are shown in Figure \ref{fig:modelscale}.

Coding data tuning consistently enhances LLMs' reasoning capacities across different model families as their size scales up to 13B. This reaffirms the conclusion drawn from smaller model comparisons: coding data can effectively improve LLMs' reasoning capacities during IFT, helping LLMs align better with user needs and handle complex logical intents. The improvement trends from coding data are similar between 13B models and their corresponding 7B models, highlighting the potential of coding data to enhance instruction fine-tuned models on reasoning tasks for larger models.

\noindent\textbf{Tuning with more coding data can better improve the overall reasoning capacities of LLMs.} To emphasize the impact of coding data proportions on overall reasoning capacities, we introduce another IFT dataset with 25\% coding and 75\% natural text, tune Llama-1 and Llama-2 models (7B and 13B parameters) on it, and evaluate each model on all reasoning tasks. Performance trends across 0\%, 25\%, 50\%, and 100\% coding data are shown in Figure \ref{fig:modelspropotion}.

 Across different model families and scales, the overall trend of gradual improvement persists after introducing the new 25\% coding data setting, reinforcing the findings from Table \ref{sec:overall} that coding data plays a key role in evoking the reasoning capabilities of pretrained LLMs. We also observe that performance at 50\% coding data is not always better than at 25\%, and vice versa, likely due to variance among backbones.

\begin{table}
\centering
\vspace{-5mm}
\caption{\textbf{Response format transition statistics.} Count of instances where the \texttt{General}-tuned model provides incorrect answers in either text or code format (upper header) and the \texttt{Code}-tuned model corrects them (lower header), using Llama-1-13B and Llama-2-13B backbones.}
\vspace{-4mm}
\resizebox{\linewidth}{!}{
\small
\begin{tabular}{l cccc |c}
\toprule
\texttt{General Output} & \multicolumn{2}{c}{text} & \multicolumn{2}{c|}{code} & \multirow{2}{*}{total} \\
\texttt{Code Output} & text & code & text & code \\ 
\midrule
Llama-1 & 1778 & 359 & 83 & 67 & 2287 \\
Llama-2 & 2591 & 224 & 63 & 49 & 2927 \\
\bottomrule  
\end{tabular}
}
\vspace{-5mm}
\label{tab:overalltransition}
\end{table}

\begin{figure*}[t!]
\centering
\vspace{-2mm}
\includegraphics[width=0.85\linewidth]{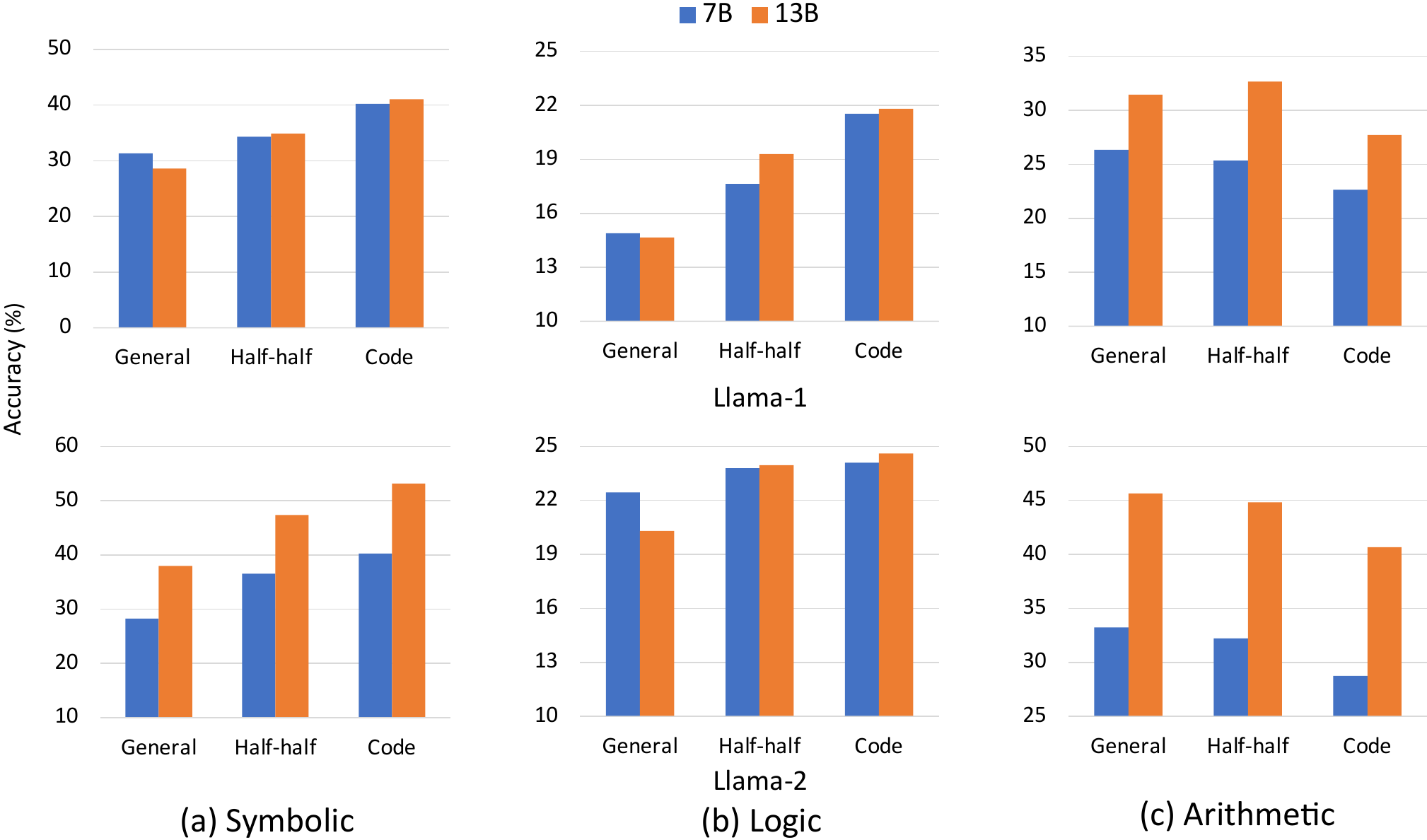}
\vspace{-2mm}
\caption{\textbf{Results comparison under each reasoning domain across different model families.}} 
\label{fig: subdomain}
\vspace{-5mm}
\end{figure*}

\noindent\textbf{Coding data tuning enhances reasoning in natural text responses.}
To investigate if coding data tuning helps solve these reasoning tasks by merely relying on producing better code, we examine the presence of code in each response under the condition where the model tuned on \texttt{Code} corrects the wrong outcomes of the model tuned on \texttt{General}. We employ GPT-3.5-turbo to detect the presence of code in responses and count the number of different answer format transitions (e.g., \texttt{General} outputs text and \texttt{Code} outputs code). The results with Llama-1-13B and Llama-2-13B as backbone are shown in Table \ref{tab:overalltransition}. The prompt for determining the response format is in Table \ref{tab:prompt-response-format} in the Appendix.

The \texttt{Code}-tuned model \emph{primarily uses pure text responses} to correct answers where the \texttt{General}-tuned model is wrong across both model families instead of relying solely on programming skills. Specifically, \texttt{Code}-tuned models successfully resolve \( \frac{1778 + 83}{2287} = 81.4\% \) of questions in Llama-1 and \( \frac{2591 + 63}{2927} = 90.7\% \) of questions in Llama-2 using natural text responses. Additionally, \texttt{Code}-tuned models do not always use the same answer format as the \texttt{General}-tuned model. They automatically choose different formats to obtain correct answers in \( \frac{359 + 83}{2287} = 19.3\% \) and \( \frac{224 + 63}{2927} = 9.8\% \) of cases using Llama-1 and Llama-2, respectively. These results further illustrate that coding data tuning successfully elicits logical thinking in LLMs beyond just in-domain skills, enabling them to answer complex questions in proper formats and improve reasoning task performance.

\vspace{-2mm}
\subsection{Results of Different Reasoning Domains}
\vspace{-1mm}
Previous discussions have focused on overall performance across reasoning tasks, but distinct domains require specific skills. For instance, symbolic reasoning tasks like letter concatenation need tokenization skills, which benefit more easily from coding data tuning, while logical reasoning requires multi-hop contextual analysis, and arithmetic tasks need enhanced quantitative reasoning \citep{yue2023mammoth}. To pinpoint the capabilities and limitations of coding data tuning across different domains, we analyze per-domain performance using Llama-1 \citep{touvron2023llama} and Llama-2 \citep{touvron2023llama-2} across various model sizes. Average results across four datasets per domain are shown in Figure \ref{fig: subdomain}.

\noindent\textbf{Coding data affects each reasoning ability differently.}
We observe distinct performance patterns in symbolic, logical, and arithmetic reasoning tasks. For \textbf{symbolic reasoning} in Figure \ref{fig: subdomain} (a), the performance improvement from \texttt{General} to \texttt{Code} tuning is significant and steadily increases. This highlights the effectiveness of code-specific data tuning in enhancing the models' foundational reasoning ability. In \textbf{logical reasoning} tasks shown in Figure \ref{fig: subdomain} (b), while all models benefit from more coding data in the IFT dataset, the gains from increasing coding data from 50\% to 100\% are minor compared to the jump from 0\% to 50\%. This diminishing return indicates that while coding data enhances reasoning skills, its utility is limited for tasks requiring advanced cognitive functions. For \textbf{Arithmetic reasoning} (Figure \ref{fig: subdomain} (c)): The best performance is seen with either \texttt{General} or \texttt{Half-half}, likely due to the need for models to understand diverse intentions in real-world applications \citep{cobbe2021training}. Tuning with a diverse IFT dataset better meets these needs than coding data alone. Despite this, the performance of \texttt{Code}-tuned models remains appealing given the limited data diversity, emphasizing the importance of coding data for LLM tuning.

\begin{figure*}[t!]
\centering
\includegraphics[width=0.85\textwidth]{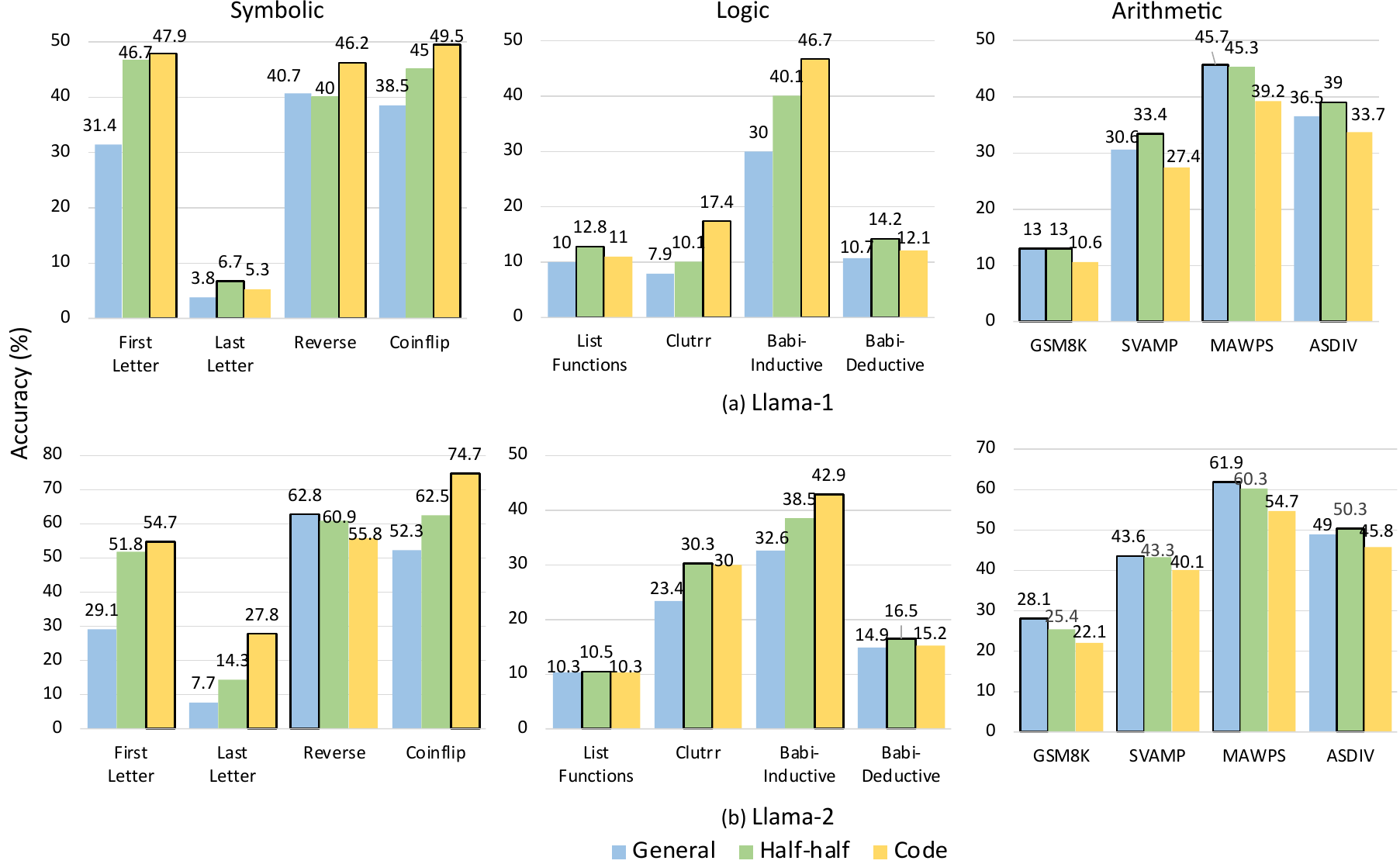}
\vspace{-3mm}
\caption{ \textbf{Results comparison for each dataset across on  (a) Llama-1-13B and (b) Llama-2-13B.} The best results of each dataset are highlighted with a black frame. } 
\vspace{-5mm}
\label{fig: eachdataset}
\end{figure*}

\noindent\textbf{Within each reasoning domain, models demonstrate similar performance trends.} 
The performance trend within each reasoning domain is consistent across model families, despite variations in architecture and pretraining datasets. This alignment suggests that the underlying factors for eliciting pretrained models' capabilities could be similar within each domain. Our analysis also shows that larger models follow the same trend as their smaller counterparts within the same family, indicating that scalability does not disrupt the effect of coding data during IFT. These findings underscore the potential for the transferability of coding data's effect across various LLM backbones and scales at the tuning stage.

\begin{table}[h!]
\centering
\vspace{-2mm}
  \caption{\textbf{Proportions (\%) of response transitions across reasoning domains.} Transition types from \texttt{General}-tuned to \texttt{Code}-tuned models, represented as `\texttt{General} (text/code) $\to$ \texttt{Code} (text/code)', across different reasoning domains for Llama-1-13B and Llama-2-13B.}
  \vspace{-2mm}
  \resizebox{0.85\linewidth}{!}{
 \Huge
    \begin{tabular}{l cccc }
     \toprule
      & \multicolumn{4}{c}{Llama-1}  \\
   & text $\to$ text & text $\to$ code  & code $\to$  text & code $\to$  code   \\
   Symbolic &  58.9 & 30.4 &  4.8 & 5.9 \\
   Logic &  93.2 & 4.8 &  0.9 &1.1 \\
   Arithmetic & 88.0 & 6.5 &  4.9 &0.6 \\
    \midrule
    & \multicolumn{4}{c}{Llama-2} \\
       & text $\to$ text & text $\to$ code  & code $\to$  text & code $\to$  code   \\
    Symbolic &  85.5 & 10.1 &  1.1 & 3.3 \\
     Logic&  96.4 & 3.0 & 0.5 & 0.1  \\
 Arithmetic&  85.1& 8.6 &  0.8 & 5.5 \\
     \bottomrule  
    \end{tabular}
  }
 \label{tab:propotion}
 \vspace{-5mm}
\end{table}

\noindent\textbf{\texttt{Code}-tuned models prefer enhancing reasoning with pure text when the \texttt{General}-tuned models output pure text, but preferences diverge across domains when they output coding.}
We further investigated response transitions where the \texttt{Code}-tuned model corrects errors from the \texttt{General}-tuned model. Table \ref{tab:propotion} shows the proportions of transition types from \texttt{General} to \texttt{Code} for each domain, using Llama-1-13B and Llama-2-13B as backbones.

We find that for both model families, the proportions of text $\to$ text' transitions are consistently and significantly higher than those of text $\to$ code' across different domains. This reinforces the finding in section \ref{sec:overall} that coding data tuning enhances reasoning task performance by genuinely improving reasoning abilities rather than relying solely on in-domain programming skills. Conversely, when \texttt{General}-tuned models produce incorrect answers involving coding data, \texttt{Code}-tuned models do not show a consistent preference for one format over the other. Specifically, code $\to$ text' is preferred in 2 out of 6 settings, while code $\to$ code' is preferred in 4. These results demonstrate that \texttt{Code}-tuned models can automatically apply appropriate skills to successfully answer questions based on domain properties.

\vspace{-2mm}
\subsection{Task-specific Reasoning Capabilities Analysis}
\vspace{-1mm}
The previous subsection highlights domain-level similarities and divergences, which may become more complex at the task level. To investigate this, we delve into each dataset to explore how coding data impacts task-specific reasoning capabilities. Results for models tuned using Llama-1-13B and Llama-2-13B backbones are presented in Figure \ref{fig: eachdataset}.

\noindent\textbf{Coding data benefits task-specific abilities in Llama-1 and Llama-2 comparatively in symbolic and logic reasoning, but diverges in arithmetic. }
We observe that models fine-tuned on datasets incorporating coding (\texttt{Half-half} and \texttt{Code}) demonstrate similar levels of superiority on task-specific abilities across different model families in symbolic and logic reasoning. Specifically, in symbolic tasks, these models either outperform or match their counterparts, \texttt{General}, in all four tasks on the Llama-1 and three out of four tasks on the Llama-2, respectively. For logical reasoning, the inclusion of coding data benefits all tasks in both model families.  However, the impact of coding data on arithmetic reasoning presents more variation. In the Llama-1, the incorporation of coding data yields better or comparable results across all tasks. Conversely, this benefit is observed only in SVAMP and ASDIV tasks for Llama-2.  These results indicate that the benefits of coding data on different LLM backbones are more similar than different, with variations emerging based on the nature of tasks within each subdomain.

\begin{figure*}[t!]
\centering
\vspace{-3mm}
\includegraphics[width=0.95\textwidth]{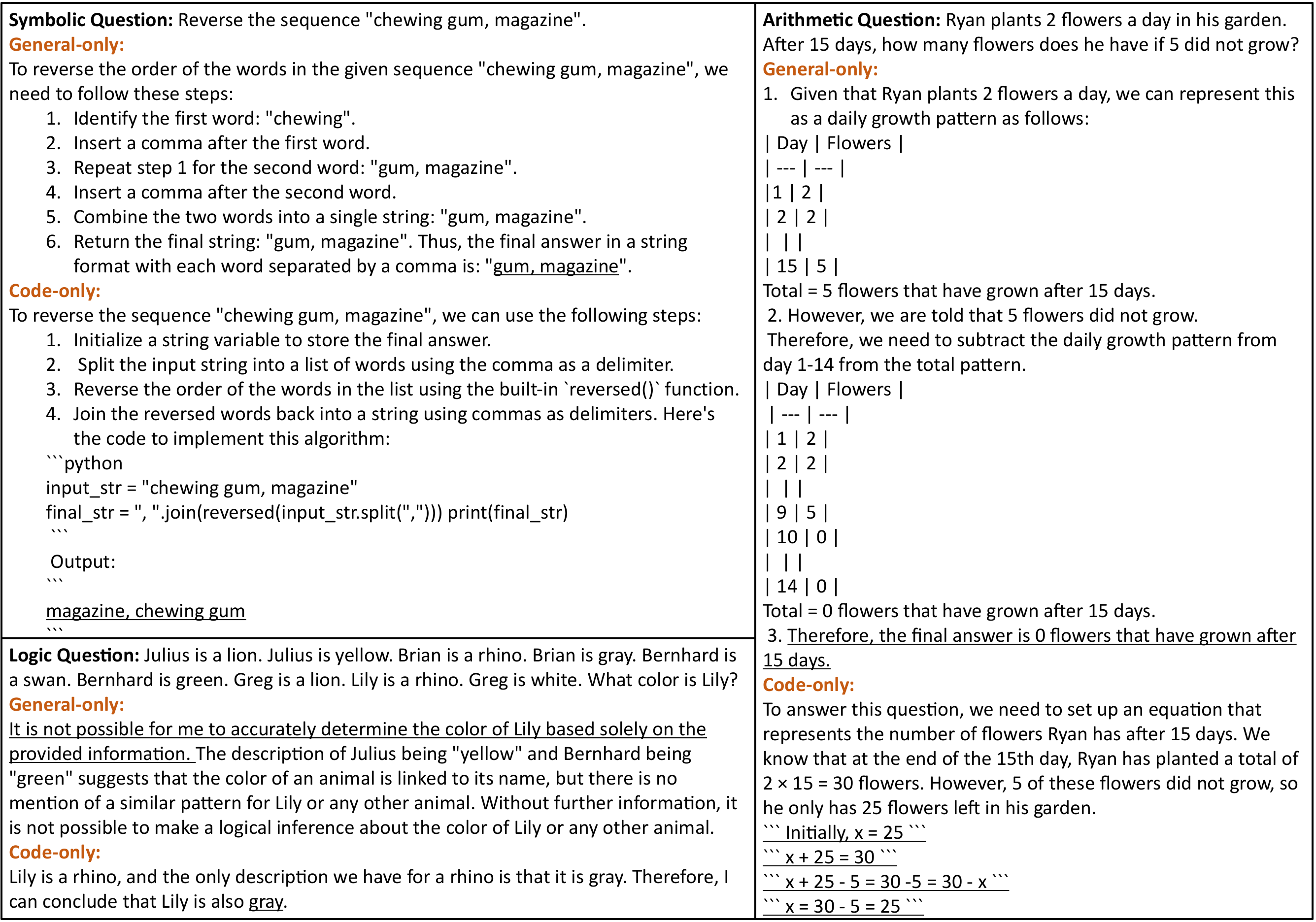}

\caption{\textbf{Case study on each reasoning domain (not cherry-picked).} We underline \underline{the final answer} of each model for clarity. } 
\vspace{-6mm}
\label{fig: casestudy}
\end{figure*}

\noindent\textbf{The optimal proportion and impact of code vary across tasks.}
We examine the impact of coding data proportions on task performance across different model families, finding no single proportion consistently superior for enhancing task-specific reasoning abilities. For Llama-1, the \texttt{Half-half} setting achieves or matches the best performance in eight tasks, while four tasks excel with the \texttt{Code} setting. For Llama-2, \texttt{Half-half} and \texttt{Code} perform best in five and four tasks, respectively. Although \texttt{Code}-tuned models don't outperform \texttt{Half-half}-tuned models in the number of tasks, they often yield greater improvements in specific tasks. For instance, in Llama-1, the \texttt{Code} setting shows a significant gain of 16.7 in the Babi-Inductive task, compared to the 3.5 gain with \texttt{Half-half} in Babi-Deductive. Similarly, in Llama-2, the \texttt{Code} setting achieves an impressive gain of 25.6 in the First Letter task, far exceeding the 7.0 gain with \texttt{Half-half} in Babi-Deductive. These findings emphasize the strategic consideration of coding proportions to optimize model effectiveness for specific goals during IFT.

\noindent\textbf{The majority of optimal coding data proportions are consistent across model families.} When investigating tasks that benefit from coding data in both model families, we find consistent optimal strategies for most tasks. Specifically, out of 8 tasks analyzed, 5 exhibit the same optimal coding data proportions across model families, while 3 require different strategies. This suggests that the proportion of coding data in IFT datasets can similarly enhance task-specific reasoning abilities across different model families, indicating a foundational influence of coding data that is generally model-agnostic. However, variability in the remaining tasks underscores the need for flexible adaptation strategies for each model family.

\vspace{-2mm}
\subsection{Case Study} 
\vspace{-1mm}
To show how coding data tuning corrects responses for different reasoning tasks, we focus on instances where the \texttt{Code}-tuned model succeeds while the \texttt{General}-tuned model fails, using Llama-1-13B as the backbone. We randomly select one instance from each reasoning domain, as shown in Figure \ref{fig: casestudy}.

For the symbolic task, the \texttt{Code}-tuned model uses Python code to correctly reverse the sequence, while the \texttt{General}-tuned model, confused by phrase construction, fails to execute the reverse action, resulting in an incorrect answer. In the logic task, the \texttt{General}-tuned model fails to follow the inductive reasoning path required for the correct answer. Conversely, the \texttt{Code}-tuned model delivers a concise and accurate answer, demonstrating improved logical capacity due to coding data tuning. For the arithmetic question, both models use structural outputs to answer, but the \texttt{General}-tuned model generates an incorrect structural reasoning path, resulting in a wrong answer. Conversely, the \texttt{Code}-tuned model leverages its in-domain coding capacity by outputting structural math equations to obtain the correct final answer. These case studies showcase how coding data tuning enhances LLM reasoning capacities to properly answer diverse questions.

\vspace{-2mm}
\section{Conclusion} \label{sec:conclusion}
\vspace{-1mm}
We studied the impact of coding data on LLMs' reasoning capacities during the IFT stage using a multi-perspective approach. Resource limitations are prevented by examining larger models like Llama-2/3 70B and we conduct evaluations focused on generative benchmark tasks within three reasoning domains. We highlighted the general effects of coding data tuning, suggesting that future research should explore detailed aspects such as diverse coding data types, content formats, and low-quality coding data. We hope this work inspires further research on LLMs' reasoning, including instruction fine-tuning, evaluation, and analysis.

\section*{Acknowledgments}
We gratefully acknowledge financial support by the National Institutes for Health (NIH) grant NIH 7R01HL149670.

\bibliography{custom}

\clearpage
\newpage

\appendix

\section{Prompt Detail} \label{subsec:prompt}
\subsection{Open API prompts}
\vspace{-3mm}
\begin{table}[h!]
\caption{Prompt for data classification} \label{tab:prompt-classfication}
\vspace{-3mm}
    \centering
    \small
    \begin{tabular}{p{7.8cm}}
        \toprule
\textbf{system prompt:} \\
You are an annotation expert tasked with categorizing conversations between humans and AI. Review each conversation and assign it to one of these categories: "Math", "Coding", or "Others".\\
Use the following guidelines:\\
Math: Assign this category if the conversation focuses on mathematical problems or concepts.\\
Coding: Choose this category for conversations that involve actual coding.\\
Others: Use this category for conversations that do not clearly fit into "Math" or "Coding," or are only slightly related to these topics.\\
\\
For generating output:\\
1. If necessary, include your reasoning within 150 words for the category selection BEFORE the final answer.\\
2. Your response  MUST contain the chosen category, formatted as: [[Category]]. \\ 
\\
For example, if a conversation is about solving a calculus problem, your response would be:\\
"Since the conversation is centered around solving a mathematical problem, it falls under the Math category. [[Math]]"\\
\\
\textbf{conversation prompt:}\\
Human:\\
\{human\_value\}\\
\\
AI:\\
\{ai\_value\}\\
\bottomrule
\end{tabular}
\end{table}
\vspace{-3mm}
\begin{table}[h!]
\caption{Prompt for coding data diversity analysis.} \label{tab:prompt-datadiversity} 
\vspace{-3mm}
    \centering
    \small
    \begin{tabular}{p{7.8cm}}
        \toprule
\textbf{system prompt:} \\
You are an annotation expert tasked with categorizing conversations between humans and AI. Review each conversation related to coding and assign it to the appropriate code category, such as Python, JavaScript, Java, C++, C\#, Ruby, Swift, Go, Kotlin, R, SQL, PHP, etc.\\
\\
For generating output:\\
1. If necessary, include your reasoning within 50 words for the category selection BEFORE the final answer.\\
2. Your response MUST contain the chosen category, formatted as: [[Category]]. For example, if a conversation is about C++, your response would be: "Since the conversation involves C++ code, it falls under the C++ category. [[C++]]"\\
3. If you cannot categorize a conversation into a specific class, output [[Others]].\\
4. You MUST try your best to label the conversation into a specific class instead of directly using [[Others]].\\
\\
\textbf{conversation prompt:}\\
Human:\\
\{human\_value\}\\
\\
AI:\\
\{ai\_value\}\\
\bottomrule
\end{tabular}
\end{table}

\begin{table}[h!]
\caption{Prompt for coding data detection in response.} \label{tab:prompt-response-format} 
\vspace{-3mm}
    \centering
    \small
    \begin{tabular}{p{7.8cm}}
        \toprule
        
\textbf{system prompt:} \\
You are a helpful AI assistant tasked with classifying whether the given input contains code or only general natural text. If code exists, output `1'; otherwise, output `0'. Ensure to only output `1' or `0' without any thinking path.\\
\\
\textbf{content prompt:}\\
Input: \{model generated response\}\\
Output:\\
\bottomrule
\end{tabular}
\end{table}
\begin{table}[h!]
\caption{Prompt for answer extraction.} 
\label{tab:prompt-extractor}
\vspace{-3mm}
    \centering
    \small
    \begin{tabular}{p{7.8cm}}
        \toprule
\textbf{system prompt:} \\
You are a helpful AI assistant to extract the final prediction of a candidate's answer to a given question.\\
\\
Here are several requirements:\\
\\
(1) You MUST extract the final prediction in the candidate answer, e.g. a numeric value, true or false, yes or no, or other types, etc., within <prediction> XML tag.\\
(2) If you cannot find the final prediction in the candidate answer, you need to cast it as "none" and output it in an XML tag.\\
\\
\textbf{quetsion-answer pair prompt:}\\
Here is <Question, Candidate answer> pair:\\
\\
Question:\\
\{question\}\\
\\
Candidate answer:\\
\{response\}\\
\\
Please extract the final prediction in the candidate answer within  <prediction> XML tag.\\
\bottomrule
\end{tabular}
\end{table}

\subsection{IFT models prompts}

\begin{table}[h!]
\caption{Prompt for tuning instruction-tuned models.} 
\label{tab:prompt-training}
\vspace{-3mm}
    \centering
    \small
    \begin{tabular}{p{7.8cm}}
        \toprule
Human:\\
\{instruction\}
\\
AI:\\
\{response\}\\
\bottomrule
\end{tabular}
\end{table}

\begin{table}[h!]
\caption{Instructions for reasoning tasks.} 
    \centering
    \small
    \begin{tabular}{p{1.5cm} |p{6cm}}
    \toprule
   First Letter
   
   Last Letter & Provide your thought process step by step to answer the question.
   
   Question:
   
   \{question\}\\
   \midrule
   Reverse List & Given a sequence of words, reverse the order and output a new string. Provide your thought process step by step.
   
   Question:
   
   \{question\}\\
   \midrule
Coin Flip& 
   A coin can be either flipped or not flipped by people. Each flip changes the side of the coin to its opposite. Answer the question step by step.
   Question:

   \{question\}\\
   \midrule
   List Functions &  Given several input-output pairs, provide your thought process step by step to determine the function that generates the outputs from the inputs and then use this function to generate an output list for a given input.\\
   \midrule
Clutrr & Provide your thought process step by step to answer the question of the relationship between two individuals based on the description. The possible relationships include aunt, son-in-law, grandfather, brother, sister, father, mother, grandmother, uncle, daughter-in-law, grandson, granddaughter, father-in-law, mother-in-law, nephew, son, daughter, niece, husband, wife, and sister-in-law.'

Description:

\{story\}
 
Question:

What is \{name 2\}'s relationship to \{name 1\} ? \\
   \midrule
   Babi\-Inductive

   Babi\-Deductive
   
   & Provide your thought process step by step to answer the question based on the description.
   
   Description:
   
   \{passage\}

   Question:
   
   \{question\}\\
      \midrule
    GSM8K
    
    SVAMP 
    
    MAWPS 
    
    ASDIV
   &Answering a math question with your thought process step by step.
   
   Question: 
   
   \{question\}\\
\bottomrule
\label{tab:prompt-generation}
\end{tabular}
\end{table}

\newpage

\section{Coding Data Diversity Analysis}
\label{subsec: code_diversity}
\begin{figure}[h!]
\centering
\includegraphics[width=0.3\textwidth]{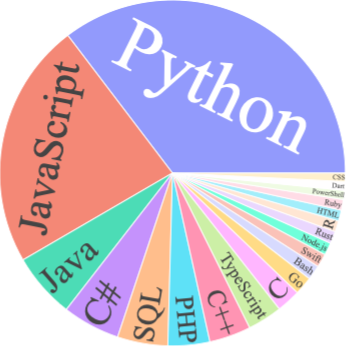}
\caption{ The distribution of the top 20 code categories diversity in the \texttt{Code} dataset.} 
\label{fig: code_divesity}
\end{figure}

As shown in Figure \ref{fig: code_divesity}, Python is the most common programming language, appearing in more than 30\% of instances in the Human-AI conversations from the ShareGPT source. JavaScript is the second most common, present in around 15\% of conversations. This phenomenon suggests that these two programming languages may have the most significant impact on the model's performance during our IFT stage, while other languages might have a much lesser influence due to their limited proportions. Future work could investigate the effects of different coding language types by considering specific structures, syntax complexities, and idiomatic usage patterns among various languages. Understanding these nuances may help clarify how these specialized characteristics in coding data can affect the model's general reasoning capacities.

\section{More About Task Description}
\label{sec: reasoningtask}
\subsection{Data statistics.} 

\begin{table}[h!]
	\centering 
   \caption{\textbf{Task data statistics.}}
  \resizebox{\linewidth}{!}{
 \small
    \begin{tabular}{lcccc}
     \toprule
    \multicolumn{5}{c}{Symbolic} \\
    \midrule
    Task& First Letter & Last Letter & Reverse List&   Coin Flip\\
    No. instances & 1500& 1500& 1000 & 600 \\
    \midrule
     \multicolumn{5}{c}{Logic} \\
        \midrule
 Task& List Functions  & Clutrr & Babi- Inductive & Babi-Deductive\\
 No. instances & 1000 &2000 & 1000 & 1000 \\
         \midrule
      \multicolumn{5}{c}{Arithmetic} \\
        \midrule
 Task& GSM8K & SVAMP & MAWPS & ASDIV\\
 No. instances & 1319 & 1000 & 2065 & 2167 \\
     \bottomrule  
    \end{tabular}
      }
 \label{tab:stat}
\end{table}

\subsection{Discussion on reasoning task selection.} 

In our work, we conduct experiments across three reasoning domains: \textit{symbolic}, \textit{logical}, and \textit{arithmetic}. We chose these domains because they primarily evaluate reasoning capacities without relying on intensive knowledge in areas such as medicine, law, or finance. This allows us to focus on how coding data affects pure logical reasoning, rather than on how it elicits specific knowledge from pretrained LLMs. Additionally, we carefully select these domains to comprehensively evaluate the reasoning capacities of LLMs from various angles. Specifically, symbolic manipulation involves tasks that are straightforward for humans but can challenge models in fundamental reasoning \citep{Wei2022ChainOT}. Logical reasoning, central to human intelligence, combines inductive and deductive processes. It requires deep contextual understanding and abstract thinking to extrapolate principles from limited observations and generalize across diverse situations \citep{saparov2024testing,qiu2023phenomenal}. Lastly, arithmetic reasoning is essential for evaluating LLMs' ability to execute complex multi-hop and quantitative reasoning tasks, which are crucial for real-world applications \citep{yue2023mammoth, liang2023holistic}.

\subsection{Details about synthetic data generation.} 

We generate symbolic datasets following the method described in \cite{fortes2023simple}. Specifically:
\begin{itemize}
    \item \textbf{Coin Flip}: We gather data for three conditions where the number of flips equals 2, 3, and 4. Each condition contains 500 instances, resulting in a total of 1500 instances for testing.
    \item \textbf{First and Last Letters Concatenation}: Three conditions are considered, where we concatenate the corresponding first or last letters with a string containing 2, 3, or 4 words. Each condition contains 500 instances, leading to 1500 instances used for testing in both the first and last letters concatenation tasks.
    \item \textbf{Reverse Sequence}: Five conditions are tested by reversing a sequence containing 2, 3, 4, 5, or 6 phrases in a string. Each condition has 200 test cases, resulting in a total of 1000 instances in this test set. 
\end{itemize}
\vspace{-2mm}
\section{Hyperparameter Setup} \label{sec:hyp}

We show the hyperparameter settings for each backbone model used for fine-tuning. We follow Alpaca \citep{alpaca} to set up the hyperparameters for Llama-1 \citep{touvron2023llama}, Llama-2 \citep{touvron2023llama-2}, Llama-3 \citep{llama3modelcard}, Qwen-1.5 \citep{qwen}, Gemma \citep{team2024gemma}, and Mammoth \citep{yue2023mammoth} for Mistral-v0.1 \citep{jiang2023mistral}. The details are shown in Table \ref{tab:hyper}.
 
\begin{table}[h!]
\centering
  \caption{
   \textbf{Hyperparameter setup.} 
  }
  \resizebox{\linewidth}{!}{
    \begin{tabular}{c ccccc}
     \toprule
   Model Family &  Model Size &  GPUs  & Epoch & LR& Batch Size\\
    \midrule
   \multirow{2}{*}{Llama-1} &7B &  4 40G A100 & 3 & 2e-5 & 128\\
    &13B &  4 80G A100 & 5 & 1e-5 & 128\\
    \midrule
    \multirow{2}{*}{Llama-2} &7B &  4 40G A100 & 3 & 2e-5 & 128\\
    &13B &  4 80G A100 & 5 & 1e-5 & 128\\
        \midrule
    Llama-3 &8B &  4 40G A100 & 3 & 2e-5 & 128\\
    Mistral-v0.1 &7B & 4 40G A100 & 2 & 5e-6 & 128\\
    
    Qwen-1.5 &7B &  4 40G A100 & 3 & 2e-5 & 128\\
    Gemma &7B &  4 40G A100 & 3 & 2e-5 & 128\\
     \bottomrule  
    \end{tabular}
  }
\label{tab:hyper}
\end{table}

% \section{Data and Code} \label{app: data}
% We provide codebase public available in link \url{https://anonymous.4open.science/r/LLM-IFT-Reasoning-0602/}.

\end{document}